\documentclass[journal]{IEEEtran}
\usepackage{times}
\usepackage{soul}
\usepackage{xurl}
\usepackage[hidelinks]{hyperref}
\usepackage[utf8]{inputenc}
\usepackage[small]{caption}
\usepackage{subcaption}
\usepackage{graphicx}
\usepackage{amsmath}
\usepackage{amsthm}
\usepackage{amsfonts}
\usepackage{booktabs}
\usepackage{algorithm}
\usepackage{algorithmic}
\usepackage{fancyhdr}
\usepackage{array}
\usepackage{verbatim}
\usepackage[shortlabels]{enumitem}
\newcommand\copyrighttext{%
\begin{footnotesize}
  This article has been accepted for publication in IEEE Transactions on Games. This is the author's version which has not been fully edited and content may change prior to final publication. DOI 10.1109/TG.2021.3138561 \\
  \textcopyright 2021 IEEE. Personal use of this material is permitted, but republication or redistribution requires IEEE permission. See \url{https://www.ieee.org/publications/rights/index.html}.
\end{footnotesize}
}

\begin{document}

\title{Visualising Multiplayer Game Spaces}
%
%
%

\author{James~Goodman,~\IEEEmembership{Student Member,~IEEE,}
        Diego~Perez-Liebana, 
        and~Simon~Lucas,~\IEEEmembership{Senior~Member,~IEEE}}

%
%

\markboth{}
{}
%


\maketitle
\copyrighttext
\\


\begin{abstract}
We compare four different `game-spaces' in terms of their usefulness in characterising multi-player tabletop games, with a particular interest in any underlying change to a game's characteristics as the number of players changes. 
In each case we take a 16-dimensional feature space, and reduce it to a 2-dimensional visualizable landscape. 

We find that a space obtained from optimization of parameters in Monte Carlo Tree Search (MCTS) is the most directly interpretable to characterise our set of games in terms of the relative importance of imperfect information, adversarial opponents and reward sparsity. These results do not correlate with a space defined using attributes of the game-tree.

This dimensionality reduction does not show any general effect as the number of players. We therefore consider the question using the original features to classify the games into two sets; those for which the characteristics of the game changes significantly as the number of players changes, and those for which there is no such effect.
\end{abstract}


%
\IEEEpeerreviewmaketitle

\section{Introduction}

Features within the game-tree such as branching factor, stochasticity and state-space size have classically been used to grade games in terms of the challenges that specific AI and search algorithms are capable of addressing. For example, the challenge of Go has been expressed in terms of its $10^{171}$ possible states, or branching factor of about 180, while an RTS game such as Starcraft ups these numbers to around $10^{1685}$ and $10^{50}$ respectively~\cite{Ontanon_2017}.
In the different context of game design, Elias et al.~\cite{elias2012} use 32 different characteristics that vary from statistics such as the number of players, the length of a game, branching factor and levels of hidden information, to more subjective measures of player interaction, positional asymmetry, sensory feedback and friendliness to spectators.

In this work we compare four approaches to characterise multiplayer games, and ask what different information these approaches provide. Each approach summarises a game as a set of 16 statistics: one is based on classic game-tree considerations; two on the relative performance of differently configured Monte Carlo Tree Search (MCTS) agents, and one on the results of optimising MCTS parameters for the game.

We use eight games in the Tabletop Games Framework (TAG) as a test set~\cite{TAG_2020}. All of these are popular published games, six were developed in the last twenty years and several have won board game awards. As such these aim to be representative of tabletop games that are currently played.
These games are distinct from classic games such as Chess or Go in supporting more than two players, often with imperfect information and stochastic elements.

Our first motivation for investigating game-spaces is to consider gaps in the games currently in TAG and focus development effort in these areas. A vision for TAG is to be comprehensive across types of modern board-game, and this analysis is one step to achieving this goal.

Our second motivation is to find similarities or clustering between games to explain why specific algorithmic variants work better on some games than others and predict which will work well on new, untried games.
This can contribute to a hyper-heuristic in game-playing~\cite{Mendes_Togelius_Nealen_2016}, and/or to a more qualitative understanding of the game landscape. For example several GVGAI games can be clustered loosely into `easy', `impossible', and `susceptible to MCTS' sets~\cite{Bontrager_Khalifa_Mendes_Togelius_2016}.
Both these examples are from single-player games, for which there is no equivalent concept to the `player count' and `opponent' dimensions that are central to our multiplayer focus.

A third motivation is to see if there are significant differences in the characteristics of games for different numbers of players and for different opponents; a game where one's opponents are playing randomly is a different environment to one in which they are using MCTS.
Two-player games have been a foundation of algorithmic research, from Chess and Go to Starcraft. These games, especially when zero-sum, have a number of well-defined theoretical properties centred around the importance of Nash Equilibria and the Minimax theorem~\cite{neumann1928theorie,nash1951}.

These game-theoretical properties do not extend beyond the two player case; in a three player zero-sum game playing a Nash Equilibrium does not guarantee non-exploitability if your opponents co-operate. 
Multi-player games, by which we explicitly mean games with at least three players, are in this sense more chaotic and less prone to analytical guarantees. 
They are also often played by humans because of these interactional dynamics~\cite{Aleknevicus_2003}, and in the form of multi-agent economic or other simulations can be of direct interest in policy formation~\cite{tesfatsion2006}.

In contrast to this potential change in interactivity beyond two players, an observation of modern board games is that some of them are prone to being `multiplayer solitaire', in which there is little interaction between players so that each player can focus on maximising their own score without needing to spend time considering the impact the actions of others may have on their strategy~\cite{BGA_2020,Burgun_2015}.
For these games one would expect there to be no fundamental change with the number of players, and games may lie on a continuum from `highly adversarial' to `multiplayer solitaire'.
We hypothesize that for some games changing the number of players fundamentally changes the nature of the game in terms of the algorithmic approaches that work, and we seek to quantify this effect.
The specific research questions we ask are:
\begin{enumerate}[topsep=3pt, label=RQ\arabic*]
\item Do the four spaces share common underlying structure, or do they provide different types of information?
\item Is there clustering within spaces by specific game, player count or type of opponent, and which is most significant?
\item Do any of the game-spaces cluster or characterise games in a way that can be used in a hyper-heuristic model to predict good algorithms to use on a new, similar, game?
\item Can we classify games into `interactive/adversarial' and `multiplayer solitaire' groups?
\end{enumerate}

Our main contribution while addressing these questions is a comparative analysis of several multiplayer games using four different measures across multiple player counts and opponents. This is one of the largest and most diverse studies of modern multiplayer games with imperfect information.
One of the sets of statistics we use is based on independent optimization runs over MCTS parameters. This novel method of characterising games turns out to be the most directly interpretable (RQ1), and is a second contribution. 

\section{Background}\label{sect:background}
\subsection{Tabletop Games Framework (TAG)}
TAG is a framework for modern Euro-style board and card-games~\cite{TAG_2020}\footnote{https://github.com/GAIGResearch/TabletopGames}. These games are of research interest because of their high popularity, because they often have high levels of hidden information, stochasticity and  support more than two players~\cite{Woods_2012}. 
They contrast with 2-player, perfect information classical games such as Chess or Go, which are well covered in the Ludii project~\cite{ludii}.
TAG currently supports 10 games, of which 8 are used in this work. The remaining two games do not fit our criteria being either purely co-operative (Pandemic) or 2-player only (Tic-Tac-Toe).
The games are listed below, for detailed summaries see \url{https://boardgamegeek.com}:
\begin{itemize}[nosep]
    \item{Dots and Boxes (1889). A perfect information game in which player take turns to connect adjacent dots on a 9x9 grid. A player scores 1 point for each grid square completely surrounded by their lines.}
    \item{Love Letter (2012). A game of role deduction with 16 cards. Each turn a player draws a card and discards one to target another player, potentially knocking one or both out of the current round. The highest surviving card wins.}
    \item{Uno (1971). Players match suits and numbers to discard all cards from their hand before their opponents. Points are scored for the value of cards held by opponents.}
    \item{Virus (2015). Players play cards from their hand to construct a healthy body of four organs (cards), and use Virus cards to infect the organs of other players.}
    \item{Exploding Kittens (2015). Players draw cards and are knocked out if they draw an Exploding Kitten. Cards can be used to peek at and manipulate the draw deck, to affect the chance of other players Exploding.}
    \item{Colt Express (2014). Players plan a (partially observable) sequence of actions to rob a train. These plans are executed in a second phase, and actions in the plan need to be executed in the light of the current situation, not necessarily the one anticipated when planning.}
    \item{Diamant (2005). A simultaneous-move push-your-luck game. Each turn players decide whether to continue exploring a mine. Treasure can be found on each turn, or the mine may collapse. The fewer players in the mine, the more treasure each gets, but they lose it all on collapse.}
    \item{Dominion (2008). A Deck-building card game in which a player needs to build an `engine' (a set of cards with complementary functions), and then use this to gain victory point cards. Each game has ten card types in play; the experiments here use the recommended first game set.}
\end{itemize}

\subsection{Search in multiplayer games}
Extending classic minimax search from 2-player to multiplayer games in the 1980s revealed new problems to solve even with perfect information.
Luckhart and Irani~\cite{Luckhart_Irani_1986} introduced the Max$^N$ algorithm in minimax search, with each agent optimising their own value. This introduces two new issues in multiplayer games; the need for more explicit opponent models and the inability to use vanilla $\alpha\beta$-pruning~\cite{Korf_1991}.

Some pruning methods applicable in multiplayer games were later developed, but these do not prune as much of the game-tree as 2-player $\alpha\beta$-pruning, and using different opponent models further reduces pruning options~\cite{Carmel_Markovitch_1997,Sturtevant_Korf_2000}.
Another approach was to use a Paranoid search that converts a multiplayer game into a 2-player zero-sum game by assuming that all the other players are acting to minimise our score, and hence enabling $\alpha\beta$-pruning~\cite{Sturtevant_Korf_2000}. 
Best Reply Search (BRS) is a relaxation of pure paranoia, with only the opponent that can damage us most making a move each turn~\cite{Schadd_Winands_2011}.
Paranoid search performs better than Max$N$ in 4-player Chess~\cite{Lorenz_Tscheuschner_2006}, and BRS and Paranoid search are found to work better than MCTS in several multiplayer games for small computational budgets~\cite{Baier_Kaisers_2020}. 
The Paranoid and Max$^N$ methods were adapted to MCTS, and Max$^N$ found to be better than the paranoid approach in 3-player Chinese Checkers~\cite{Sturtevant_2008}.

Opponent Move Abstraction (OMA) shares information between MCTS tree nodes reached by the same set of agent moves, reducing the contribution of opponent moves~\cite{Baier_Kaisers_2020}. 
The idea is that incorporating every opponent move in the search tree can lead to very shallow search, and it may be better to devote the computational budget to searching deeper in the tree based on the moves we can make.

The opponent modelling requirement arises because when considering an opponent's move in response to ours we need to model their model of us. Only in a 2-player game with rational players is the use of a single minimax tree alternating between players theoretically valid~\cite{Luckhart_Irani_1986}. 
Several works address opponent modelling in multiplayer search, including the implications of different opponent utility functions~\cite{Sturtevant_Zinkevich_Bowling_2006,Sturtevant_2008,Donkers_2003,Carmel_Markovitch_1996}. A detailed review is avoided here because opponent modelling is not a focus in this work. 
Inn all our experiments, opponent actions are either modelled in the search tree or as random actions. The same utility function is used for all players (and opponent models) in each game.

All of the works cited previously in this section use perfect information versions of games, and seven of the eight games used here have hidden information of some form. This makes classic search difficult to use without converting them to perfect information variants as in \cite{Sturtevant_2008}. 
MCTS can be used more easily in stochastic games with imperfect information using Open Loop and Information Set MCTS methods as described in the next section.

\subsection{Monte Carlo Tree Search (MCTS)}

MCTS has been used in many games~\cite{Coulom_2006,Chaslot_De_Jong_Saito_Uiterwijk_2006,Browne_MCTS_2012}. It is an anytime algorithm that uses a time budget to search the forward game tree. On each iteration four steps are followed:
\begin{enumerate}
\item{Selection. Select an action to take from the current state. If all actions have been selected at least once then the best one is picked using the Upper Confidence for Trees (UCT) equation \cite{UCT_2006}:
$J(a) = Q(a) + K\sqrt{\frac{\log(N)}{n(a)}}$
The action $a$ with largest $J(a)$ is selected. $N$ is the total number of visits to (iterations through) the state; $n(a)$ is the number of those visits that then took action $a$; $Q(a)$ is the mean score for all visits to the state that took action $a$; $K$ controls the trade-off between exploitation, and exploration choosing actions with few visits so far. This step is repeated down the tree of game states until a state is reached with previously untried actions.}
\item{ Expansion. Pick one of the untried actions at random, and expand this, creating a new state in the game tree.}
\item{Rollout. From the expanded state, take random actions for a number of steps (or game-end) to obtain a score.}
\item{Back-propagation. Back-propagate this score up the tree. Each state records the mean score of taking a given action as $Q(a)$ that will affect future Selection steps.}
\end{enumerate}
Once the time budget has been used the action at the root state with the highest score or most visits is executed.
Since our interest is in multiplayer games, we include a number of variations to the core MCTS algorithm. The nine parameters we configure are summarised in Table~\ref{tab:MCTSParams}:

\begin{table} [t]
\centering
\begin{tabular}{p{1.6cm}>{\raggedright\arraybackslash}p{2.4cm}>{\raggedright\arraybackslash}p{3.25cm}}
\toprule
Parameter  & Values & Description \\
\midrule
Tree & \{UCB & Standard UCT formula\\
Policy  & Alpha & The AlphaZero formula\\
  & EXP3 & Softmax Selection \\
   & RM\} &  Regret Matching UCT \\
\midrule
Opponent & \{MaxN & Maximise their score\\
Tree  & Paranoid & Minimise our score\\
 Policy & SelfOnly\} & Act randomly\\
\midrule
 Final & \{Robust & Most visited action\\ 
 Policy & Simple\} & Most valuable action\\
\midrule
 TreeDepth & \{1, 3, 10, 30, 100\} & Max Depth of Tree\\ 
\midrule
 Rollout & \{0, 3, 10, 30, 100\} & Max actions per rollout \\
\midrule
 Redeterminise & \{false, true\} & Information Set MCTS\\
\midrule
 Open Loop & \{false, true\} & Use Forward model in tree\\
\midrule
 K & \{0.01, 0.1, 1, 10, 100\} & Exploration coefficient for UCB and Alpha\\
\midrule
 explore $\epsilon$ & \{0.01, 0.03, 0.1, 0.3\} & Exploration chance for EXP3 and Regret Matching\\
\bottomrule
\end{tabular}
\caption{MCTS Parameters included in optimization run. See main text for details on each.}
\label{tab:MCTSParams}
\end{table}

\begin{itemize}
\item Tree Policy. As well as the UCT policy~\cite{UCT_2006} we try a variant inspired by AlphaZero~\cite{AlphaGo_2017} that modifies the exploration term to $\sqrt{N}/(1+n(a))$,
and EXP3 and Regret Matching (RM) policies described in \cite{Lisy_2014} that have better theoretical behaviour in adversarial environments.
\item Opponent Tree Policy. We consider three variants to model the opponent in the tree. `MaxN' assumes that each other player maximises their utility (we assume all players use the same utility function, described in Methods \ref{sect:method}) and tracks all players in one tree; at each node the acting player's utility is used to make a decision. This is the Multiplayer-UCT of ~\cite{Sturtevant_2008}. `Paranoid' instead models other players as minimising our utility, and ignoring their own.
`SelfOnly' tracks just the agent's own actions in the tree, and models other players as moving randomly. 
\item Final Policy. This controls the decision at the root node after search. `Robust' picks the action with the most visits; `Simple' picks the one with the highest mean value. 
\item Max Tree Depth. The maximum depth to which the tree will be constructed before a rollout starts.
\item Rollout Length. The maximum length of each rollout. This many actions will be taken in a rollout before the final state is evaluated (or the game terminates).
\item Redeterminise. If {\tt true} then Information Set MCTS is used~\cite{Cowling_Powley_Whitehouse_2012}. This redeterminises the game state at the start of each iteration of MCTS, reshuffling across all possible sets of the hidden information.
If {\tt false}, Perfect Information MCTS is used with a single shuffling of the hidden information used for all MCTS iterations. 
\item Open Loop. When {\tt true} the forward model advances the game state from the root on every iteration, and the underlying state may be different on each node visit, due to a stochastic environment or actions of other players~\cite{Perez_Dieskau_Hunermund_Mostaghim_Lucas_2015}.  When {\tt false} the forward model is not applied in the tree, and each leaf stores the state that caused its expansion. This reduces the number of forward model calls but does not account for stochastic environments.
\item K. The exploration weighting in the UCT formula.
\item $\epsilon$. K is not used in EXP3 and RM selection tree policies. Instead a random action is taken with probability $\epsilon$ 
\end{itemize}

\subsection{NTBEA}
The N-Tuple Bandit Evolutionary Algorithm (NTBEA) is described in~\cite{Lucas_Liu_Perez-Liebana_2018}. It has been benchmarked against other optimisation algorithms in stochastic game environments and proven to be good at finding a good set of parameter settings within a fixed computational budget \cite{Lucas_Liu_Bravi_Gaina_Woodward_Volz_Perez-Liebana_2019}. 
Any optimization algorithm suitable for stochastic environments could be used, and it is not an essential part of our contribution. 

\subsection{Dimensionality Reduction}
A variety of dimensionality reduction techniques can reduce data in $\mathbb{R}^n$ to $\mathbb{R}^m$ with $m < n$~\cite{van2009dimensionality}. If $m=2$ then we have the benefit of being able to visually plot the data.
We use two common techniques, PCA and t-SNE. 

Principal Components Analysis (PCA) finds the linear subspace $\mathbb{R}^m$ within $\mathbb{R}^n$ that preserves as much of the variance in the original data as possible when points in $\mathbb{R}^n$ are mapped to the closest point in the subspace~\cite{hotelling1933analysis}. 
This linearity means that each feature axis in $\mathbb{R}^n$ is mapped to a straight line in $\mathbb{R}^m$. 
The `loading' of a feature in the original space onto a dimension in the reduced space is the correlation of this line to the reduced dimensional axis. This is always between -1 and +1.
This is an advantage of PCA as the axis in the reduced space can be interpreted in terms of the original features.  
The axes of the best $\mathbb{R}^2$ linear subspace can be freely rotated without changing the subspace itself. The `rotated components' are the axes in the reduced space that maximise the sum of the squared loadings. This does not change the space, but may make it more interpretable.

PCA cannot represent non-linear manifolds in the data, even if these are low-dimensional. It also preserves relative long-range relationships between points at the expense of short-range ones, which can be a disadvantage in some cases.

t-SNE is a non-linear reduction method for high-dimensional datasets that instead focuses on having points that are close in $\mathbb{R}^m$ close in $\mathbb{R}^2$, and is agnostic about where distant points are placed. It aims to show distinctly separated clusters, but will not show patterns at greater distance~\cite{Maaten_Hinton_2008}. 
In practice we found that t-SNE plots evenly spread the points across the available space with little interpretable structure, and the results are not included. 

We use Parallel Analysis in PCA to estimate the number of significant components in the data~\cite{hayton2004factor}. This finds the number of principal components with larger eigenvalues than expected by chance; purely random data would give zero significant components. 
A `scree plot' plots the eigenvalues in descending order of magnitude with a second line for the eigenvalues for purely random data. The point at which the first line drops to or below the level of the second is the cut-off for significance.

Canonical Correlations Analysis (CCA) is a dimensional reduction technique to test if two different representations of the same underlying objects have a common underlying structure.
CCA takes representations in two separate spaces ($x$, $y$) of the same set of objects, and finds linear projections of both into a common lower dimensional space ($z_x$, $z_y$) that maximises the covariance between the two sets of projected points~\cite{barber2012}. 
As with PCA, each feature in the original $x$ and $y$ spaces has a `loading' into $z \in \mathbb{R}^2$, allowing comparisons between the features from the two original spaces.

\section{Method}\label{sect:method}
Four sets of statistics are used to compare the eight games, each with a total 
of 16 features/dimensions.
\begin{enumerate}
\item Statistics from the game tree, such as size of action space, reward sparsity and hidden information. We do not expand the whole game-tree and collect these statistics only for the parts of the tree visited during play. 
\item Statistics from the optimisation of agents to play each game. For each game independent NTBEA runs generate a distribution over parameter space. The marginal distributions then give a 16-dimensional space.
\item The win rates of 16 different agents in each game against fixed opponents. 
\item The win rates of the same 16 agents in a Round Robin tournament against each other.
\end{enumerate}

Player counts of 2, 3 and 4 are used for each game, and three fixed opponents: a random player (RND), a one-step lookahead player (OSLA), and a simple MCTS player with a budget per decision of 20ms. The MCTS player uses Open Loop Paranoid UCT with rollout length and tree depth of 10.
This generates 9 data points for each game (or 3 for the Round Robin statistics).

Each set of statistics is designed to convey the same amount of information. We are interested in which sets of statistics are useful to differentiate games from each other. Each combination of game, player count and opponent is embedded in each $\mathbb{R}^{16}$ space, and PCA used to reduce the dimensionality for visualization. 
The following questions are asked:
\begin{itemize}
\item Is clustering greater by game, the number of players, or the type of opponent? For each space 
we conduct a Mann-Whitney U test on the distances between points for the same game against a null hypothesis that these have the same distribution as all inter-point distances. This tests if points from the same game are closer together (clustered) or dispersed. We repeat this for points of the same player count, and for the same opponent. Results in~\ref{sect:clustering}.
\item  What is the underlying number of dimensions (significant components) in each space using parallel analysis? Results in Section~\ref{sect:clustering}.
\item For each space what is the interpretation of the significant components. Does this, or the visualizations, provide insight into game clusters, or which algorithms work best with which game features? Results in~\ref{sect:GAResults} to \ref{sect:RRresults}.
\item Does Canonical Correlations Analysis (CCA) show connections between the spaces. Results in \ref{sect:CCAResults}. This excludes the Round Robin space as it has 24 points, versus 72 in the others and cannot be compared with CCA.
\end{itemize}

When a set of games is run, the position of each agent is randomised so that the order of decision-making varies across games. This may be important if there is an advantage to going first for example. 
The TAG framework controls the main game state, and passes a copy of this to each agent when a decision is required, with all hidden information randomised. The agent returns its action selection, and the copy of the game state is discarded. This avoids information sharing between agents.

The same utility function is used by all non-RND players in all games, with +1 for a win, -1 for a loss, 0 for a tie. Where a non-terminal state is evaluated the game score is used, scaled to be within a [0, +1] range, with +1 corresponding to a theoretical maximum score (the details vary by game as they have different scoring mechanics).

\subsection{Game Attributes}\label{sect:GameAttr}
For each game, player count and agent type (RND, OSLA, Simple MCTS) we run 1000 games and track 16 statistics. 
In each game all the players use the same agent type.
The statistics are:
\begin{itemize}
    \item The mean times to copy a game state and apply the forward model to a game state. This gives two statistics related loosely to game complexity.
    \item The mean number of decisions taken in each game, and the coefficient of variation (CoV) measured over one game. CoV is used as it is dimensionless and comparable across games with different numbers of decisions.
    \item The count of different components of the game state (e.g. the number of cards, board positions, loot tokens, and any game component that can change state in a game). We measure the starting value, the mean value over the game, and the CoV over one game. This gives three statistics related to the size of the state space.
    \item The proportion of the game components hidden from the player. We measure the starting proportion, the mean over the game, and the CoV.
    \item The mean size of the action space (the number of actions available), the CoV of this over the game, the mean maximum action space in a game, and the skew of the action space distribution.
    \item The mean score for a player over a game, and the CoV. All games provide a score at each turn to each player bounded to the range [-1, +1], with the extremes indicating actual win or loss. A draw scores zero.
\end{itemize}

We use Action Space size and not Branching Factor due to the restriction of gathering statistics during play; we cannot enumerate all values that a revealed card might have to calculate how many different states may result from an action.
None of the games have dice, or any stochasticity outside deck shuffling. 
Random outcomes, and any difference between Action Space and Branching Factor, is due to hidden information in unknown card or other values.

\subsection{NTBEA Optimization}\label{sect:NTBEA}

\begin{figure}[t]
\centering
\includegraphics[width=0.5\textwidth]{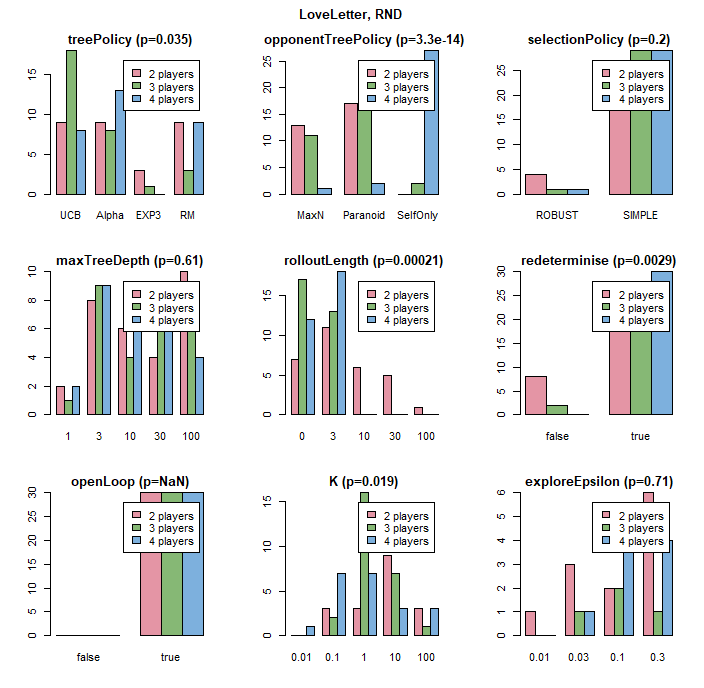}
\caption{\small Fingerprint of Love Letter from 30 independent optimisations (against a random opponent). Each facet displays the marginal distribution over a different parameter. For Love Letter the optimal opponent tree policy (p = 3.3$\times10^{-14}$) and rollout length (p = 2$\times10^{-4}$) both change significantly with the player count under a chi-squared test of homogeneity. The Open Loop parameter is vitally important as all 30 runs select $true$, while performance is less sensitive to $maxTreeDepth$ with flatter distribution.}
\label{fig:fingerprintLL}
\end{figure}

For each combination of game, player count and opponent we run 30 independent NTBEA optimisations over MCTS parameters to produce a `fingerprint' as described in~\cite{Goodman_2021}.
Figure~\ref{fig:fingerprintLL} shows an example. 
A marginal distribution over a parameter provides more information than a single optimised value;
a flat distribution shows insensitivity to a parameter, while a sharply peaked distribution shows it is vitally important.
We optimise an MCTS agent to win against the fixed opponent with a 40ms computational budget per decision. The objective optimised scores +1 for a win, -1 for a loss and 0 for a tie where several players jointly win. The game score is not used in this optimisation objective.
Experiments are run on a 2.6 GHz Intel Xeon Gold 6240 CPU. 

The dimensions and values optimised are in Table~\ref{tab:MCTSParams}. 
The search space has nine dimensions and a size of 48,000, so each run can sample from about 10\% of the space. We do not seek to run each optimization run to convergence as we are also interested in how fast each parameter converges to generate the marginal distribution.

The marginal distributions of seven of the parameters (excluding K and $\epsilon$) were used to calculate the 16 statistics. 
Because the sum of all settings of a parameter must equal the total number of NTBEA runs (30), each dimension contributes a number of statistics equal to the total number of possible settings for the dimension, minus one.
For example, one statistic is the number of NTBEA runs that recommend Open Loop as {\tt true}, and we do not have another statistic for the {\tt false} setting as this would be redundant.

\subsection{Relative Agent Performance}\label{sect:relative}
We use the win rates of 16 different agents as the features in this space.
For each agent we run 1000 games for each player count, each game, and each opponent (RND, OSLA, Simple MCTS).
Twelve of these agents are chosen from high-scoring MCTS parameters generated by NTBEA optimisations, selecting for a broad range of different parameter settings shown in Table~\ref{tab:sampleAgents}.
The remaining four agents are the RND and OSLA agents, plus two Random Mutation Hill Climbing (RMHC) algorithms with horizons of 3 and 20 actions. These mutate one action at random in each iteration, and return the first action in the best performing plan.

In the 1000 games in each set, one random player uses the current agent and all the other players use the same opponent type.
This set of statistics is expected to give the same broad types of information about games as the NTBEA statistics, as both derive from the performance of different MCTS configurations.

\begin{table}[t]
    \centering
    \small
    \begin{tabular}{cccccccccc}
    \toprule
        Ref & Roll & OL & OppT. & Tree & K & $\epsilon$ & D & IS  \\
         \midrule
        A & 3 & Y & MaxN & Alph & 0.1 &  & 1 &   \\
        B & 10 & Y & P'noid & Alph & 0.01 & & 100 & Y  \\
        C & 0 &  & MaxN & E3 &  & 0.01 & 1 & N  \\
        D & 100 & Y & P'noid & E3 &  & 0.1 & 3 & Y  \\
        E & 30 &  & MaxN & RM &  & 0.3 & 30 & N  \\
        F & 0 & Y & P'noid & RM & & 0.3 & 10 &   \\
        G & 0 &  & MaxN & UCT & 0.01 & & 100 &   \\
        H & 10 &  & Self & UCT & 0.1 &  & 30 &  \\
        I & 3 & Y & Self & UCT & 10 &  & 1 & Y \\
        J & 10 & Y & P'noid & UCT & 1 &  & 3 & Y \\
        K & 30 & Y & Self & E3 & & 0.03 & 10 & N \\
        L & 10 & Y & Self & RM & & 0.03 & 3 & Y \\
         \bottomrule
    \end{tabular}
    \caption{The 12 MCTS agents used for the Agent Performance statistics. Columns correspond to the parameters in Table~\ref{tab:MCTSParams}. 
    D is the tree-depth, OL is Open Loop and IS is Information Set MCTS.}
    \label{tab:sampleAgents}
\end{table}

\subsection{Round Robin Tournament}\label{sect:roundrobin}

For each game and player count a round robin tournament was run using the 16 agents from \ref{sect:relative}. Agents were randomly matched, with no agent duplicated in any one game, and each agent played 10,000 games.
This gives a single data point for each game and player count. 
No analysis by opponent is possible in this data set, as results for each agent are against a uniform mixture of the other fifteen. 
This has an advantage in that these are on average better opponents than the RND, OSLA and simple MCTS players used in the previous experiments.
Hence we expect this may provide better information on `hard' games in which almost any MCTS agent can beat a RND agent.

    \begin{figure}[t]
      \centering
    \includegraphics[width=0.5\textwidth]{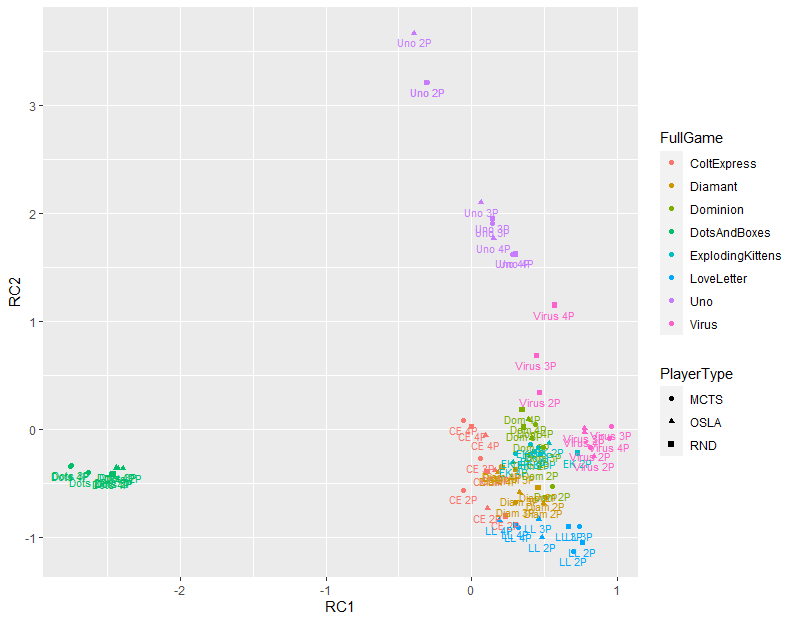}
    \caption{PCA projection of the full 16D Game Attribute space to 2D.}
    \label{fig:GameAttributePCA}
    \end{figure}

\section{Results}\label{sect:results}

\subsection{Clustering and effective dimension}\label{sect:clustering}

The results show that varying the opponent or player count does not have a significant effect on the clustering of points. Points for a given game are on average clustered closely together in all four of the game spaces.

\begin{table}[t]
    \centering
    \begin{tabular}{lcccc}
    \toprule
        Statistics & Dim & Game & Players & Opponent \\
        \midrule
        Game Attr& 16 & 2e-16 & 0.83 & 0.93 \\
        Game Attr& PCA-2 & 2e-16 & 0.70 & 0.25 \\
        \midrule  
        NTBEA  & 16 & 2e-16 & 0.78 & 0.52 \\
        NTBEA & PCA-2 & 2e-16 & 0.68 & 0.66 \\
              \midrule
        Agent Perf & 16 & 2e-16 & 0.52 & 0.20 \\
        Agent Perf& PCA-2 & 2e-16 & 0.46 & 0.39 \\
                \midrule
        Agents RR & 16 & 1e-3 & 0.66 & - \\
        Agents RR & PCA-2 & 4e-10 & 0.54 & -\\
        \bottomrule
    \end{tabular}
    \caption{Clustering by Game, Player Count or Opponent. The entries are the p-values from a Mann-Whitney test on point-to-point distances. Low values indicate evidence of clustering by that attribute.}
    \label{tab:wilcoxonGlobal}
\end{table}

\begin{figure*}[t]
\begin{subfigure}{.49\textwidth}
  \centering
\includegraphics[width=\textwidth]{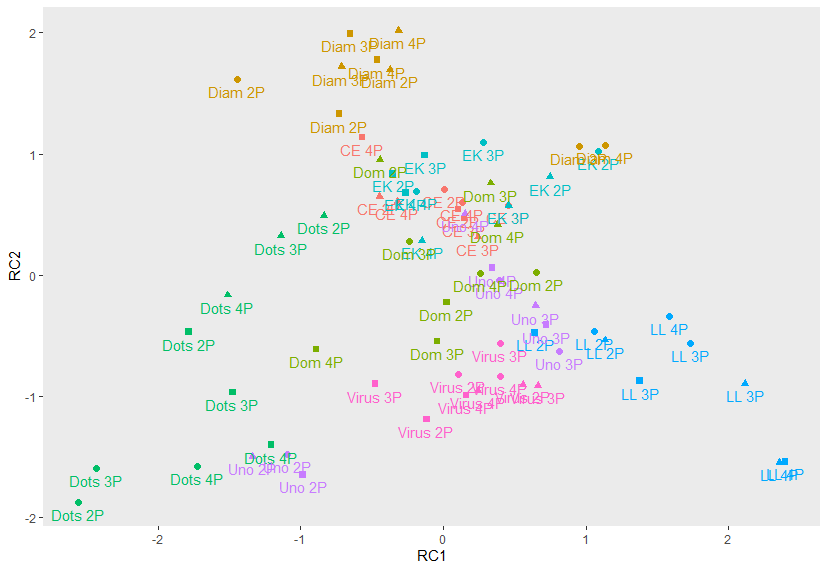}
\end{subfigure}
\hfill
\begin{subfigure}{.49\textwidth}
  \centering
\includegraphics[width=\textwidth]{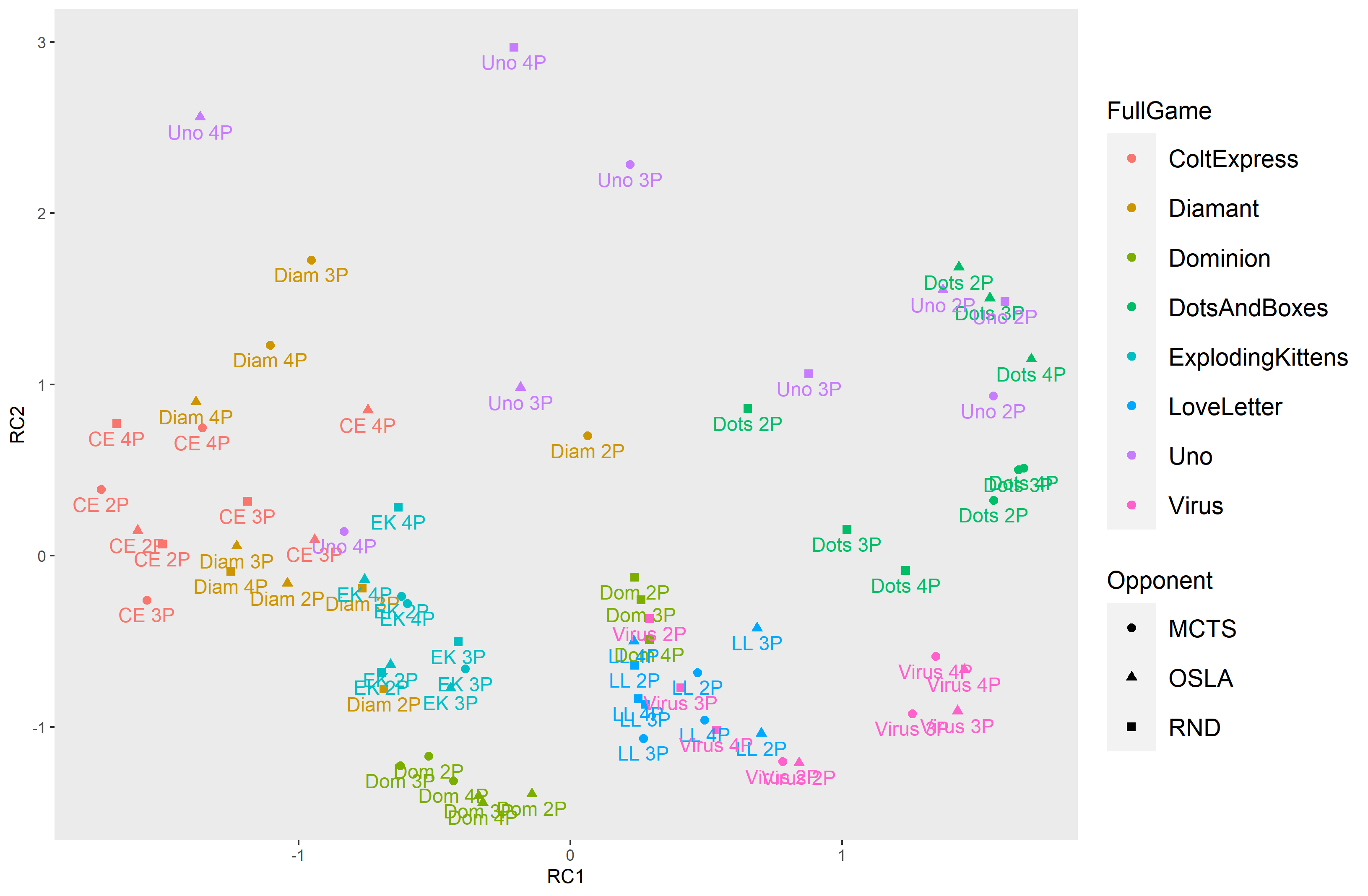}
\end{subfigure}
\caption{The left-hand side uses PCA to project the full 16D NTBEA space to 2D. The right-hand side does the same to the data from Agent Performance. (The RC1 and RC2 axes are not comparable between the two diagrams.)}
\label{fig:NTBEAPCA}
\end{figure*}

This clustering is visible in the PCA projections in Figures \ref{fig:GameAttributePCA} and \ref{fig:NTBEAPCA}.
Table~\ref{tab:wilcoxonGlobal} summarises the results of the Mann-Whitney tests on the inter-point distances for matching game/player count/opponents. 
These tests are repeated in each full-dimensional space (scaled to mean zero and unit variance) and each 2-D PCA projection. 
The results show conclusively that in all spaces points for the same game are closely clustered, regardless of player count or details or the algorithms used as opponents. The p-value from the test is $<2\times10^{-16}$ in most cases; the exception being 
the Agent Round Robin space, which has fewer points to compare, but still has $p \le 0.001$.
There is no evidence that clustering occurs by player count or opponent when measured across all eight games.

\begin{figure}[t]
    \centering
    \begin{subfigure}{0.5\textwidth}
      \includegraphics[width=\textwidth]{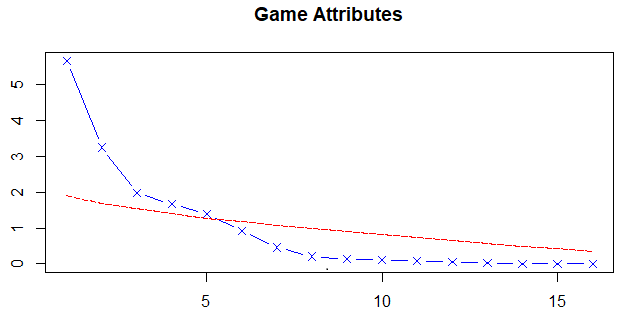}
    \end{subfigure}
    \hfill
        \begin{subfigure}{0.5\textwidth}
      \includegraphics[width=\textwidth]{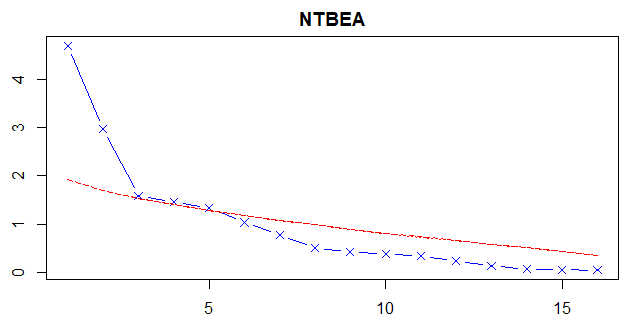}
    \end{subfigure}
    \caption{Scree diagrams show the magnitude of the eigenvalues (y-axis) of the principal components (x-axis) for Game Attribute and NTBEA spaces. The red line is the eigenvalue for random data. The blue line shows the actual eigenvalues. Where the two cross is an estimate of the underlying dimensionality. }
    \label{fig:PCAScree}
\end{figure}

Parallel Analysis shows that the Game attribute space has five significant components, with eigenvalues greater than expected by chance.
Each of the other three spaces has two significant components.
Figure~\ref{fig:PCAScree} shows the scree plots used for this analysis for the Game Attribute and NTBEA spaces. The scree plots for the Agent Performance and Round Robin spaces are similar to that for NTBEA, but with a sharper decline in significance after the first two eigenvalues.

\subsection{Game Attributes}\label{sect:GAResults}

    \begin{table}[t]
        \centering
        \begin{tabular}{lll}
        \toprule
             &   Loadings (2D) & Loading (5D) \\
             \midrule
         RC1  &  +0.92 HI (Mean) & +0.95 HI (Mean) \\
                &  +0.94 HI (Start) & +0.87 HI (Start) \\
             &  -0.91 AS (Mean) & -0.95 AS (Mean) \\
             &  -0.87 AS (Max) & -0.94 AS (Max) \\
                      &  -0.84 Copy (Median) & -0.92 Copy (Median) \\
                      \midrule
         RC2  & +0.89 SS (Start) &  +0.84 SS (Start) \\ 
               & +0.86 SS (Mean) &  +0.89 SS (Mean) \\ 
             & +0.87 Decisions &  +0.89 Decisions \\ 
             & +0.89 Decisions ($\sigma^2$) & +0.90 Decisions ($\sigma^2$) \\ 
             \bottomrule
        \end{tabular}
        \caption{Game Attribute loadings on the first two rotated components (RC) for 2D and 5D PCA. HI is Hidden Information, SS is State Space, AS is Action Space, Copy is the time to copy a state. Features with a loading of at least 0.7 are shown.}
        \label{tab:GamePCALoading}
    \end{table}
    
\begin{figure}[ht]
    \centering
    \includegraphics[width=0.5\textwidth]{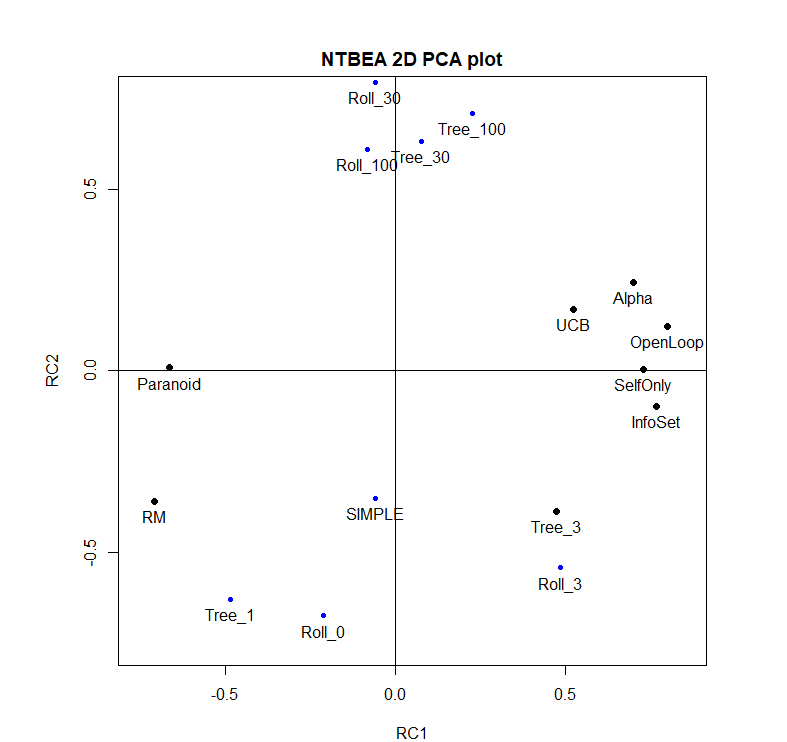}
    \caption{NTBEA 2D PCA dimension loadings} 
    \label{fig:NTBEA_PCA2D_Loadings}
\end{figure}

The results show that using the simple attributes of the game tree to characterise games is sensitive to the specific games, and is not likely to generalise well to new games.

This set of statistics has the tightest clustering by game (see Figure~\ref{fig:GameAttributePCA}), and this is relevant when interpreting the meaning of the significant components identified.
Table~\ref{tab:GamePCALoading} shows the loadings for the first 2 dimensions. 
The first component is positively correlated with the level of Hidden Information in a game, and negatively correlated with the mean Action Space, and the time to copy a state.
This seems to be purely because of the game at one extreme on this dimension in Figure~\ref{fig:GameAttributePCA}:
Dots and Boxes has zero hidden information, a slow copy function and a large action space as on each turn any two adjacent points on a five-by-five grid can be connected.

PCA finds dimensions that are good at discriminating the data by maximising variance, which will be large when many data points are at opposite extremes of a principal component.
This example shows it can be driven by differences specific to the games in the sample, and choices over implementation details.
PCA has highlighted that Dots and Boxes is the most `different' to the other games, and in what respects.

The second dimension shows a more general progression over games, and uses the length of a game (the number of decisions) and number of components in the game. Both of these are largest for Uno, a game that can extend (theoretically) forever, and 108 cards compared to Love Letter at the other extreme with a fixed 16 cards.

The three subsequent principal components find projections that put one or two games at extremes; RC3 separates out Exploding Kittens, RC4 differentiates Virus and Diamant, RC5 does the same for Dominion.
With each game forming a tight cluster in Game Attribute space, the PCA objective is maximised by separating the eight game clusters, and the results are not likely to generalise to a larger set of games.

This conclusion is confirmed if we repeat the analysis with Dots and Boxes removed. Each of the components shifts 'up' one. The current RC2 becomes RC1, and RC3 (which differentiates Exploding Kittens) becomes RC2. 
Omitting other individual games can have similar effects on other dimensions; removing Uno changes RC2 significantly and removing Exploding Kittens changes RC3.

\subsection{NTBEA statistics}\label{NTBEAResults}

The space characterises games based roughly on the requirement to consider hidden information during play, monitor adversarial opponent moves, and how long it takes for the benefit of a move to be reflected in the game score. It is stable on removal of any individual game.

Figure~\ref{fig:NTBEA_PCA2D_Loadings} shows the loading of features on the two significant dimensions in PCA.
The first component has Open Loop, Information Set and Self-Only MCTS on the positive axis, and Paranoid and Regret Matching MCTS on the negative. We expect Open Loop and Information Set MCTS to be useful in stochastic games with hidden information; while Paranoid and Regret Matching MCTS were added because of their better fit to adversarial games. 
This provides an interpretation that games on the far right of Figure~\ref{fig:NTBEAPCA} (Love Letter and Exploding Kittens) require MCTS variants that cope with high levels of hidden information.
On the left are Dots and Boxes, and 2-player Uno that require the variants designed for at adversarial environments.

The second component tracks the relative importance of deep rollouts and tree depths. Rollouts and tree depths of 30 and 100 have high positive loadings, while zero rollouts, and a tree depth of one have strong negative loadings. 
This can be interpreted as a gradation from games in which deep exploration of the game-tree is needed to get a useful reward signal (Diamant, Colt Express, Exploding Kittens), to those where looking just a small number of steps ahead suffices (Dots and Boxes, 2P Uno, Love Letter, Virus).

The points for the same game are not as tightly clustered as in the Game Attribute space (compare Figures \ref{fig:GameAttributePCA} and \ref{fig:NTBEAPCA}), with a spread of points over the space. This helps avoid the pathology noted in Section~\ref{sect:GAResults} of PCA simply differentiating one game at the extreme of each component.
The two significant components have only small changes to the loadings if the analysis is repeated with Dots and Boxes data removed, and this holds for removal of any single game from the analysis.

Figure \ref{fig:NTBEAPCA} suggests that for some games the optimal MCTS parameters vary mostly by opponent; For Dots and Boxes the three points for each of the three opponents (RND, OSLA, MCTS) are closely grouped regardless of player count, and for Dominion, the RND player is distinct. 
These are also the two games for which the RND player performs particularly badly, and is a poor opponent, while OSLA is much better than either of the others in Dots and Boxes.

For other games the points in Figure \ref{fig:NTBEAPCA} vary mostly by player count;
for Uno we see that the 2P optimal parameters are very different to those for 3 or 4 players, closer to the perfect information Dots and Boxes.

\subsection{Agent Performance and Round Robin}\label{sect:RRresults}

These spaces have the same underlying two significant components as each other, which do not map directly to the two components in the space using NTBEA statistics. 

\begin{table}[t]
    \centering
    \begin{tabular}{lllll}
    \toprule
    Feature     & RC1 & RC1-RR & RC2 & RC2-RR  \\
        \midrule
    A & +0.72 & -0.74 &  & \\
    C & +0.83 & -0.90 & & \\
    D & & +0.96 & &  \\
    E & & +0.94 & & \\
    F & & -0.75 &  & \\
    G & & -0.78 & & \\
    H & -0.72 & +0.73 & & \\
    J & -0.77 & +0.93 & & \\
    OSLA & & & +0.88 & -0.89 \\
    RMHC-3 & & & +0.80 & -0.96 \\
    RMHC-20 & & & +0.78 & -0.96 \\
    RND & & & & -0.93 \\
    \bottomrule
    \end{tabular}
    \caption{PCA Loadings $ > 0.7$ for Agent Performance and Round Robin (RR) statistics. Letter features refer to agents from Table~\ref{tab:sampleAgents}. RMHC-N is Random Hill Climbing with a horizon of N.}
    \label{tab:AgentPCALoading}
\end{table}

\begin{figure}[t]
  \centering
\includegraphics[width=0.5\textwidth]{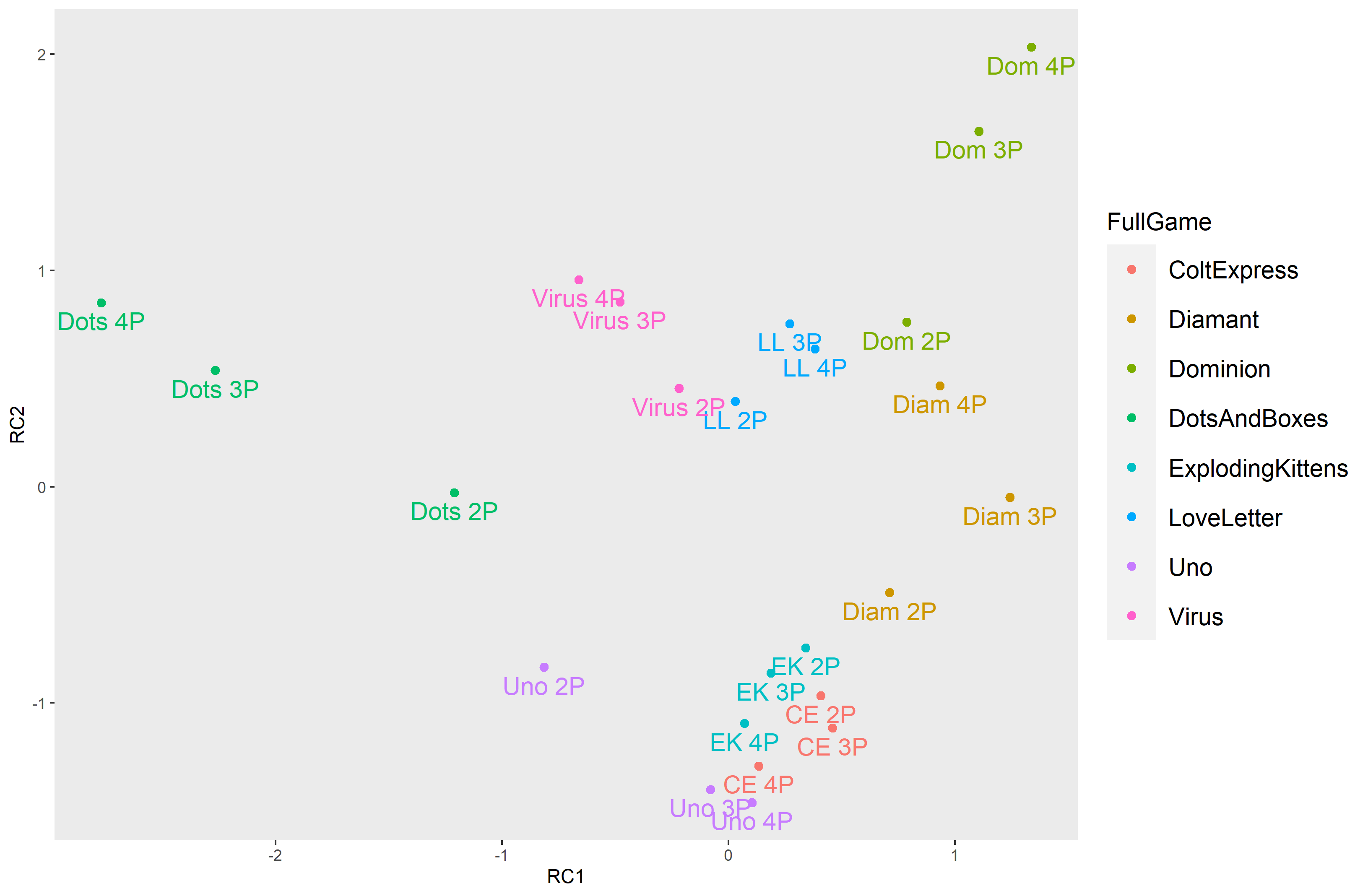}
\caption{2D PCA projection of the Round Robin tournament space.}
\label{fig:RoundRobinPCA}
\end{figure}

The two significant dimensions in both Agent Performance and Round Robin (RR) spaces are in Table~\ref{tab:AgentPCALoading}. The PCA plots are in Figures~\ref{fig:NTBEAPCA} and~\ref{fig:RoundRobinPCA}.
Table~\ref{tab:AgentPCALoading} shows the components for both spaces are fundamentally the same, with the signs on the axes flipped.
As this similarity in loadings suggests, the Round Robin PCA projection broadly clusters with similar neighbours to the Agent Performance projection, although there is not a tight 1:1 mapping.
These two sets of features represent the same information.

The first component (RC1) includes all Agents with rollout of 10 or greater on one side of the first, and those with 0 or 3 on the other.  Similarly, all the Agents except one (I) that use Information Set MCTS are on the same side.
RC1 combines elements of both the two principal components in the NTBEA optimization statistics in Figure~\ref{fig:NTBEA_PCA2D_Loadings}, but the fact that each letter represents a combination of individual MCTS settings make this much harder to interpret. 

The second component (RC2) has OSLA at one extreme, along with Agent C, which effectively \emph{is} OSLA (tree depth of one, no rollout). It also incorporates the Random Mutation Hill Climbing agents, and hence information from non-MCTS agents.
The game with the most obvious distinction between NTBEA and Agent Performance spaces in Figure~\ref{fig:NTBEAPCA} is Uno, which is an outlier on the upper edge of the Agent Performance plot. This is because RMHC and RND score better in Uno than the other games.

\begin{figure}[t]
    \centering
      \includegraphics[width=0.5\textwidth]{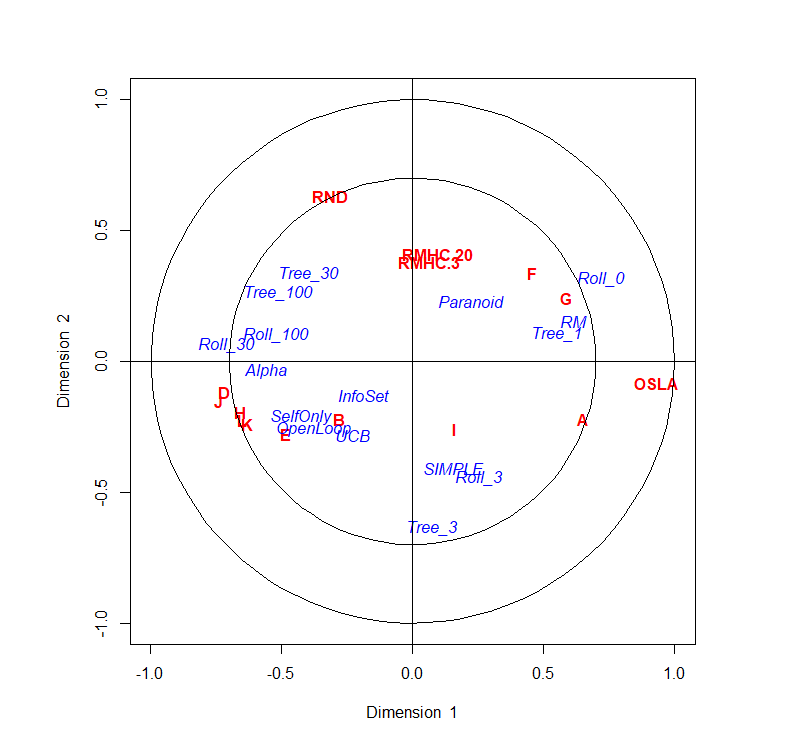}
    \caption{Plot of CCA mapping of individual features from NTBEA (red) and Agent Performance (blue) spaces. Features closer to the unit circle have a strong influence on the projection; the inner circle is drawn at the point where half the original variance in the feature is transferred. Features close to each other are correlated.}
    \label{fig:CCANTBEAPlayers}
\end{figure}

\subsection{Canonical Correlation Analysis}\label{sect:CCAResults}

The results of CCA between the Game Attribute and other spaces are not shown. 
CCA mappings using the Game Attribute space preserve the approximate clustering, but as in the PCA results in \ref{sect:GAResults} these find the projections that best preserve the Dots and Boxes separation from the other points.

Figure~\ref{fig:CCANTBEAPlayers} shows how NTBEA and Agent Performance spaces map to a joint space under CCA. 
A feature on the unit circle contributes all of its variance, while one at the centre is ignored. The main contributions are from long and short rollouts and performance of the RND and OSLA agents. 
Features close to each other in Figure~\ref{fig:CCANTBEAPlayers} are correlated. The cluster of 6 Agents (D, E, H, I, J, K) are all closely related to long rollouts, while F, G, and A are linked to short rollouts as expected. 
Hence the first dimension in the joint space separates games that need large rollouts to play well from those for which one-step-lookahead is ideal. It is similar to RC2 in the NTBEA space and RC1 in the Agent Performance space. 

The second dimension of the joint space is not so clearly attributable. It includes the performance of the RND and RMHC agents from RC1 of the Agent Performance space, with mid-range values of Tree depth and rollout length.

\section{Discussion}\label{sect:discussion}

\subsection{RQ1 - is there a common structure to the spaces?}

The Game Attribute space has five significant dimensions, and the others all have two. PCA is a more useful method than t-SNE with $\mathbb{R}^{16}$ feature spaces. t-SNE loses any long-range structure in the original and is difficult to interpret in terms of the original features.

The components of the NTBEA space over optimised MCTS parameters are interpretable as the relative importance of hidden information versus strongly adversarial opponents on one dimension, and the depth of the `forward planning horizon' needed to obtain a reward signal on the other.

The two most significant components of the Game Attribute space are potentially similar, with one strongly influenced by the amount of hidden information present, and the second by the length of the game (number of decisions taken). 
However there is no correlation between placement of games in the two spaces despite this similarity. NTBEA space is more `useful' as it provides a guide to the importance in game-play: a game may have large amounts of hidden information, but this is irrelevant if a player does not need to consider the possible different values to play well.
Only in a sub-set of games does this hidden information require specific algorithmic adaptations.
Love Letter and Exploding Kittens have high-impact events that can trigger when an unknown card is revealed (disqualifying a player from the round in Love Letter, and knocking them out the game in Exploding Kittens). We speculate that this is behind the pattern observed here. 
Unseen cards in Dominion, Colt Express and Uno can affect the result, but not to this extreme.


The second NTBEA dimension (forward planning horizon) might be expected to correlate with the second component in Game Attributes (number of decisions), as the longer a game lasts, the more benefit is to be gained from longer rollouts.
This is not the case when we compare Figures~\ref{fig:GameAttributePCA} and~\ref{fig:NTBEAPCA}. 2P Uno has the longest games (more than 700 moves), and does not benefit from long rollouts, while Diamant is the second shortest game, and benefits most from long rollouts.
Again, we have a pattern that simple analysis of the Game Attributes does not convert naturally into predictions of which algorithmic variants will be most helpful, despite initial similarity in interpretation.

If the analysis is repeated with Dots and Boxes removed, then in the Game Attribute space each of the components shifts 'up' one. The initial RC2 becomes RC1, and RC3 becomes RC2, and so on. 
If the same is done with for the NTBEA space the space does not significantly change, with the same two significant components, and this holds for removal of any single game from the analysis.
Unlike the NTBEA space, the Game Attribute space does not generalise. It differentiates one or more games at an extreme in each dimension, and the pattern depends on the games included.
It can help visualise the space to see where there are gaps in the current coverage, and decide what further games could usefully be added to TAG. A conclusion is that Dots and Boxes is an outlier as the only perfect information game, and adding games with low levels of hidden information could provide a more representative sample.

The Agent Performance and Round Robin spaces represent the same information. 
It was hoped the Round Robin space would give different information, even though the same agents were used in both. The hypothesis was that the fixed opponents may be too easy to beat, and using the other agents as opponents would mitigate overfitting to the fixed set. This does not seem to be a problem in practice with these games.

The first component of the Agent Performance space combines elements of both conponents in the NTBEA space, with the second component dependent on the performance of non-MCTS agents.
One reason for this merger of the two dimensions found in the NTBEA case is the performance of each agent is a mixture of the effect of all of parameter settings that the agent uses. The 12 MCTS agents used cannot represent every possible combination, and the space cannot disentangle the effects of individual parameters.
The NTBEA optimization approach can better disentangle such effects with features based on parameter-specific marginal distributions.
As a result the NTBEA statistics have greater representative power, are more directly interpretable, and automatically search all areas of parameter space not covered by an \emph{a priori} selection of test agents. Their downside is that they cannot include information from non-MCTS agents.


\subsection{RQ2 - is there clustering by game, player count or type of opponent?}

This has a straightforward answer from Table~\ref{tab:wilcoxonGlobal}; the game being played is more important than the number of players or opponent/player type. 
This applies across all four spaces.
The PCA plots for the NTBEA and Agent Peformance statistics show that for individual games, there can be differences by player count or oppponent type, and we follow this up in RQ4, but this does not apply consistently across all games.



\subsection{RQ3 - Can we find a hyper-heuristic model to choose algorithms on a new game?}

The two components of the NTBEA space provide a guide to setting MCTS parameters. One is interpretable as the forward planning depth needed until a useful change in score is found, and the second balances the importance of hidden information and adversarial behaviour.
However, to find the position of a new game on this map, it is first necessary to optimise MCTS parameters, in which case we already have the information we need.

An automatic hyper-heuristic would require there to be a correlation between the information in the Game Attribute space, which is fast to gather, and the information in NTBEA space, which requires more time. Since no such correlation is found, this work does not unfortunately lead directly to such a hyper-heuristic.
It is still possible to leverage domain expertise on a new game to decide on the best initial set of MCTS parameters to try. This can be a `nearest neighbour' comparison to the eight games, or a structured response to the questions:
\begin{enumerate}
    \item Does the game have a short lag between an action and a meaningful reward (change in score)? If so, use a short rollout length and tree depth to maximise the number of MCTS iterations within a budget. Longer lags will require longer rollouts.
    \item Does the value of hidden information (cards held by other players) have a major impact on your own score? If so, use Information Set MCTS.
    \item If the game has little hidden information and is highly adversarial, with opponent moves able to negatively impact one's own score, then use a Regret Matching or EXP3 selection rule instead of UCT.
\end{enumerate}

\subsection{RQ4 - Is there a distinction between adversarial and `multiplayer solitaire' games?}

The answer to RQ2 is that the position of a point in any of the spaces is determined primarily by the game, with no consistent changes for different player counts. 
The `game landscapes' can be useful for characterising games with respect to each other, but do not show a large impact from a changing player count in aggregate as hypothesized.

Figure~\ref{fig:NTBEAPCA} shows evidence that for some games at least, there is a significant change in the position of a game as the player count changes; Uno is the most obvious example.
Additionally the fingerprint of marginal distributions for Love Letter in Figure~\ref{fig:fingerprintLL} has very significant change in individual parameter distributions as the player count changes: opponent tree policy has $p=3\times10^{-14}$ for the , and rollout length has $p=2\times10^{-4}$. However for Love Letter there is no major difference is apparent in the PCA visualizations. Why not?


PCA focuses on the dimensions that differentiate the individual data points from each other in order to maximise the overall variance. Any dimensions not useful for this goal will be discarded.
In the case of Love Letter, of the 6 or so features with strong loadings on the first principal component, only one (SelfOnly) differentiates Love Letter by player count in Figure~\ref{fig:fingerprintLL}, while the others (e.g. Open Loop and Information Set MCTS) are required equally for all numbers of players.
PCA is an unsupervised technique that retains maximal information. It does not privilege those aspects of the data that differentiate by player count in one game, and the significant differences in Love Letter are lost when PCA finds the differentiating dimensions across all games.

To address the question of what changes with player count, we need to look at individual parameter changes.
To do this Fisher exact tests for every game, opponent and MCTS parameter were run on the NTBEA results to detect significant differences for different player counts. This totals 216 tests, and the Bonferroni correction was used to show p-values of less than $0.05/216$ = $2\times 10^{-4}$. Assuming each test is independent, this gives a 4.2\% chance of reporting a single significant result under the null hypothesis that there is no player count specific effect of any of the parameters. In fact 24 of the tests meet this threshold (Table~\ref{tab:playerDiff}).
This was repeated this for varying opponents and here 29 tests meet the threshold.

    \begin{table}[t]
    \centering
        \begin{tabular}{lccccccc}
        \toprule
        Game & D & OL & OT & IS & Roll & Fin & Tree  \\
        \midrule
           Colt Express &   &   &   &   & 1 &   & \\
           Diamant      &   & 1 & 1 & 1 & 1 &   & 1\\
           Dots+Boxes   &   &   & 1 &   & 2 & 1 &  \\
           Expl. Kittens&   &   & 1 & 2 &   &   & \\
           Love Letter  &   &   & 2 &   & 2 &   &   \\
           Uno          & 3 &   &   & 1 & 3 &   &  \\
             \bottomrule
        \end{tabular}
        \caption{Game/parameter combinations with significant differences in the optimised settings by player count. Each cell is the number of opponents (out of three) for which this difference was significant. D is Tree Depth, OL is Open Loop, IS is Information Set MCTS, Tree is the Tree policy (UCT/EXP3 etc.), OT is Opponents Tree policy (MaxN/SelfOnly/Paranoid) and Fin is Robust or Simple.}
        \label{tab:playerDiff}
    \end{table}

From Table~\ref{tab:playerDiff} we can say that Rollout length, Information Set MCTS and the Opponent Tree policy are important as player count changes, as was suggested in the Love Letter NTBEA fingerprint of Figure~\ref{fig:fingerprintLL}. The precise effect varies by game, with a general pattern of shorter rollout lengths and less use of a Paranoid tree policy for more than 2 players.
Of the eight games considered, Dominion, Virus and Colt Express do not show evidence of being different games based on the number of players, while the other five in Table~\ref{tab:playerDiff} do.
Conclusions from opponent differences are less general as they depend on the precise opponent used, but with RND/OSLA/MCTS opponents three games show significant changes in optimal parameters; Diamant, Dominion and Dots and Boxes.

\subsection{Limitations}
There are a number of limitations to the results.
The most obvious is that the 8 games, even when expanded to 72 environments by considering three different player counts and three different player types, fundamentally have only 8 points in Game Attribute space with no generalization to new games.
This does not apply to the other spaces 
where the nine environments per game overlap significantly. 
In the NTBEA space the first component balances the importance of hidden information with an adversarial opponent. A greater variety of games could test if this is two separate effects.

Secondly, all experiments use 40ms per decision. This permits 200-2000 MCTS iterations, varying by game, and allows 150,000 games to be run in a few days on a single CPU, but does not give a world-class play.
It is a future question how the patterns we observe across the games for MCTS parameter settings vary as we change the computational budget as in~\cite{Kim_Ashlock_2017}.

Thirdly, only MCTS parameters are used. The differences between the Round Robin and Agent Performance spaces show that using simple hill-climbing distinguishes games in ways that MCTS may not be able to. 

A fourth limitation is the use of very simple Random and OSLA opponents and a consistent utility function for opponents. In some games these agents are weak, and may not provide a good context to characterise the game, although the fact that the Agent Performance and Round Robin spaces have the same significant components suggests that this is not a general issue, even if it is for specific games. 
Additional factors exist in multiplayer games compared to 2-player games when considering utility functions and opponent modelling. Instead of a simple win/loss, an agent may value its ordinal result; is coming second of four better than coming fourth? 
Multiplayer games also introduce `kingmaking', with the actions of a third player deciding the final result of two other players; agents could have preferences over the other players.
Investigating the impact of these effects and changing the simple fixed set of opponents is a planned area of future work to address this restriction.

\section{Previous Work}\label{PreviousWork}

To review previous work on characterisation of extensive form games, we focus on work related to the relative performance of different AI algorithms. 
We split this into three categories based on the properties used to describe each game:
\begin{enumerate}
    \item Properties intrinsic to the game (player-independent)
    \item Properties of the game as it is played (player-dependent)
    \item Invert the approach by using the properties of agents that play a particular game well
\end{enumerate}

GVGAI is a common testbed because it has many games in one framework. Otherwise the work reviewed below primarily focuses on a small set of 1-4 games, and considers 1-4 features across these. These features are often determined post-hoc from a consideration of what is likely to cause algorithmic performance differences across the games.
In this work we deliberately determine the game space features a priori, as well as extending to 8 games, and 72 environments when considering different player counts and opponents.

\subsection{Intrinsic Properties}
Intrinsic properties can be calculated for a game without any need to play it well.
This advantage is offset by the computational cost of calculation when these properties rely on traversal or enumeration of the whole game-tree.
Jansen analyses complete game-trees to predict situations in which a sub-rational opponent may be exploitable. This is only possible using small random trees with up to 20 states~\cite{Jansen_1992}. 
Similar is exploration of features of games that lead to MCTS performing poorly (or well) relative to MinMax search, and the existence of `traps' in the game tree~\cite{Ramanujan_Sabharwal_Selman_2010}. These are positions from which a small number of good moves are combined with many poor ones that rapidly lead to a game loss. Finding these requires a full game-tree analysis.
Finnson and Bjornsson~\cite{Finnsson_Bjornsson_2011} use simple single-player game-trees with varying branching factor, tree depth, `traps' and monotonic progression towards an end-state 
to confirm that traps are a real problem for MCTS.


Sturtevant~\cite{Sturtevant_2008} suggests two criteria, branching factor and n-ply state variance (how much the state can change in one move). The first of these is quantifiable, the second needs a game-specific heuristic. Experiments show they are correlated with the success of MCTS variants in Chinese Checkers, Spades and Hearts; high branching factor favouring informed rollouts, and high state variance disfavouring learning heuristics such as MAST or RAVE. 
Later work considers Bias, Leaf correlation and Disambiguation Factors as features of the full game tree, and quantifies these in three card games to explain the success of Perfect Information MCTS in some games with hidden information~\cite{Long_Sturtevant_Buro_Furtak_2010}.

MCTS and Evolutionary Algorithms are tried on a number of GVGAI games in \cite{Horn_Volz_Perez-Liebana_Preuss_2016}. The authors fit a linear model to the games to quantify which attributes of games cause specific algorithms to succeed or fail. The features of games are not based on game-tree analysis and are quick to calculate. Pure MCTS is the only algorithm that deals positively with enemy NPCs, but has problems with path-finding. 

Bontrager et al.~\cite{Bontrager_Khalifa_Mendes_Togelius_2016} use a similar approach, and one most directly similar to ours. They use a set of features to describe GVGAI games, and use these to find the best algorithm for a given game. They extract 14 features from the VGDL game description and look at clusterings in this Game Attribute space, and in an Agent Performance space (1 dimension for each of the 23 algorithms). 
This shows four clusters in each space, but only low (40\%) similarity between the clusters in Performance and Feature space.

\subsection{Properties from Play}

Mendes et al.~\cite{Mendes_Togelius_Nealen_2016} use a portfolio of different algorithms to play 41 GVGAI games. Fourteen features are extracted from the game-state during play.
Various Supervised Learning methods predict the best algorithm for a held-out game. This selected the best algorithm 64\% of the time, over a 50\% baseline of picking the single best algorithm.

Other works consider game characteristics for Game generation or design~\cite{Togelius_Nelson_Liapis_2014,Browne_Maire_2010}. The latter uses 57 `aesthetic' metrics of tension and drama; how often does the lead change in a game, or the likelihood that the eventual winner was previously behind, to measure the quality of a game.
Player-specific metrics can be used to characterise games in play-testing, as in~\cite{Nelson_2011,volz20154}. They do not traverse or analyse the whole game tree, but measure attributes of the game as experienced by a given player, be that human or AI. This can focus on game-specific events, such as `Did event A happen?', or `How often did the player interact with NPC X?' This can be used to characterise different types of player and assist in GVGAI game design and testing~\cite{Guerrero-Romero_Louis_Perez-Liebana_2017}.
We differ by using general attributes applicable to any extensive form game outside the constraints of VGDL.

\subsection{Properties of Agents}

Stephenson et al.~\cite{Stephenson_Anderson_Khalifa_Levine_Renz_Togelius_Salge_2020} look at 108 GVGAI games and 27 algorithms from competition entrants, and use information theory to analyse which games discriminate best between different algorithms.
Soemers et al.~\cite{soemers2021deep} look at 15 games in Ludii played with MCTS with and without a learned heuristic function.
The score difference highlights outlier games where this performance difference metric is unusual.

Gaina et al.~\cite{Gaina_Liu_Lucas_Perez-Liebana_2017} also use GVGAI with Open Loop MCTS and RHEA variants. Two parameters of RHEA are varied and patterns found are highly game-specific with MCTS poor in Deterministic games. 
We extend this by visualising the game-space and consistently varying the player count and opponent faced (albeit with only three of each).

An empirical game theory and $\alpha$-rank approach focuses on 2-player zero-sum games, and analyses them in terms of the transitive and non-transitive relationships between sampled strategies at playing the game~\cite{Omidshafiei_et_2020,SpinningTops_2020}. 
It shares a perspective with this work of creating `tools that enable the discovery of a topology over games', and in reducing the role of human-crafted dimensions, but using graph theory to analyse the results of agent match-ups.
Features from the `response graph' of these strategic relationships are found to correlate reasonably well ($\rho\approx0.6$) with a separate measure of game complexity.
Plotting a spectral decomposition of the response graph visually groups similar games together. 

\section{Conclusion}

We have used four different sets of statistics to visualise games spaces using eight multiplayer games. 
The least generalizable is the Game Attribute space derived from game-tree statistics, for which all points for a game cluster very tightly, and the five significant components in PCA change drastically if individual games are removed.

The other three spaces all characterise games on a 2-dimensional linear subspace, although the details of this subspace differ.
The game-space defined by optimization of MCTS parameters is most interpretable because it separates the impact of each of the parameters. In comparison the spaces based on agent performance combine multiple different settings in each `feature', although they can incorporate information from non-MCTS agents.

The MCTS optimization space has two dimensions. One of these is the importance of using longer rollouts/simulations to detect a delayed reward signal (score). The second is the importance of Open Loop and Information-Set MCTS to deal with hidden information in the game, versus the importance of modelling adversarial behaviour.
Comparison with the Game Attribute space results shows no direct correlation with absolute levels of hidden information, or game length with these two dimensions. Simple measurement of attributes of the game tree cannot tell us how these impact on game-play and no automated hyper-heuristic is found to link features of the game-tree to a recommended algorithm.

In all four game-spaces the points for a given game are much more closely located than points for a given player count or opponent. We do not find evidence that modern multiplayer games in general exhibit a consistent change in character as the player count increases beyond two. 
Looking at the changes in individual MCTS parameters with player count we identify the subset of games which do exhibit a significant change for increased player count, with others that can be characterised as `multiplayer solitaire'.

The analysis of modern multiplayer games is under-explored. This work uses a new approach to analyse a diverse set of these games using the marginal distribution of optimisation results. It provides the most thorough analysis of the effect of MCTS parameterisation on them to date and is a small step towards a better understanding of the genre.

\section*{Acknowledgments}
This work was funded by the EPSRC CDT in Intelligent Games and Game Intelligence (IGGI) EP/S022325/1. 
Thanks are due 
to the anonymous reviewers, whose comments improved this paper considerably, and
to Martin Balla, Alex Dockhorn, Raluca Gaina and Raul Montoliu for contributing games to TAG!

\bibliographystyle{IEEEtran}
\bibliography{ijcai21}

\begin{thebibliography}{10}
\providecommand{\url}[1]{#1}
\csname url@samestyle\endcsname
\providecommand{\newblock}{\relax}
\providecommand{\bibinfo}[2]{#2}
\providecommand{\BIBentrySTDinterwordspacing}{\spaceskip=0pt\relax}
\providecommand{\BIBentryALTinterwordstretchfactor}{4}
\providecommand{\BIBentryALTinterwordspacing}{\spaceskip=\fontdimen2\font plus
\BIBentryALTinterwordstretchfactor\fontdimen3\font minus
  \fontdimen4\font\relax}
\providecommand{\BIBforeignlanguage}[2]{{%
\expandafter\ifx\csname l@#1\endcsname\relax
\typeout{** WARNING: IEEEtran.bst: No hyphenation pattern has been}%
\typeout{** loaded for the language `#1'. Using the pattern for}%
\typeout{** the default language instead.}%
\else
\language=\csname l@#1\endcsname
\fi
#2}}
\providecommand{\BIBdecl}{\relax}
\BIBdecl

\bibitem{Ontanon_2017}
S.~Ontanón, ``Combinatorial multi-armed bandits for real-time strategy
  games,'' \emph{Journal of Artificial Intelligence Research}, vol.~58, p.
  665–702, 2017.

\bibitem{elias2012}
G.~S. Elias, R.~Garfield, and K.~R. Gutschera, \emph{Characteristics of
  games}.\hskip 1em plus 0.5em minus 0.4em\relax MIT Press, 2012.

\bibitem{TAG_2020}
R.~D. Gaina, M.~Balla, A.~Dockhorn, R.~Montoliu, and D.~Perez-Liebana, ``Tag: A
  tabletop games framework,'' in \emph{Proceedings of the AIIDE workshop on
  Experimental AI in Games}, 2020.

\bibitem{Mendes_Togelius_Nealen_2016}
\BIBentryALTinterwordspacing
A.~Mendes, J.~Togelius, and A.~Nealen, ``Hyper-heuristic general video game
  playing,'' in \emph{2016 IEEE Conference on Computational Intelligence and
  Games (CIG)}.\hskip 1em plus 0.5em minus 0.4em\relax IEEE, Sep 2016, p.
  1–8. [Online]. Available:
  \url{http://ieeexplore.ieee.org/document/7860398/}
\BIBentrySTDinterwordspacing

\bibitem{Bontrager_Khalifa_Mendes_Togelius_2016}
P.~Bontrager, A.~Khalifa, A.~Mendes, and J.~Togelius, ``Matching games and
  algorithms for general video game playing,'' in \emph{Twelfth Artificial
  Intelligence and Interactive Digital Entertainment Conference}, 2016, p.
  122–128.

\bibitem{neumann1928theorie}
J.~v. Neumann, ``{Zur Theorie der Gesellschaftsspiele},'' \emph{Mathematische
  Annalen}, vol. 100, no.~1, pp. 295--320, 1928.

\bibitem{nash1951}
J.~Nash, ``Non-cooperative games,'' \emph{Annals of mathematics}, pp. 286--295,
  1951.

\bibitem{Aleknevicus_2003}
\BIBentryALTinterwordspacing
G.~Aleknevicus, ``Player interaction,'' 2003. [Online]. Available:
  \url{http://www.thegamesjournal.com/articles/PlayerInteraction.shtml}
\BIBentrySTDinterwordspacing

\bibitem{tesfatsion2006}
L.~Tesfatsion and K.~L. Judd, \emph{Handbook of computational economics:
  agent-based computational economics}.\hskip 1em plus 0.5em minus 0.4em\relax
  Elsevier, 2006.

\bibitem{BGA_2020}
\BIBentryALTinterwordspacing
B.~G. Atlas, ``Multiplayer solitaire,'' 2020. [Online]. Available:
  \url{https://www.boardgameatlas.com/forum/JOW1vYXlRt/multiplayer-solitaire}
\BIBentrySTDinterwordspacing

\bibitem{Burgun_2015}
\BIBentryALTinterwordspacing
K.~Burgun, ``Why many eurogames are inherently single-player games,'' Feb 2015.
  [Online]. Available:
  \url{https://keithburgun.net/why-eurogames-are-inherently-single-player-games/}
\BIBentrySTDinterwordspacing

\bibitem{Woods_2012}
S.~Woods, \emph{Eurogames: The design, culture and play of modern European
  board games}.\hskip 1em plus 0.5em minus 0.4em\relax McFarland, 2012.

\bibitem{ludii}
M.~Stephenson, E.~Piette, D.~J. Soemers, and C.~Browne, ``An overview of the
  ludii general game system,'' in \emph{2019 IEEE Conference on Games
  (CoG)}.\hskip 1em plus 0.5em minus 0.4em\relax IEEE, 2019, pp. 1--2.

\bibitem{Luckhart_Irani_1986}
C.~Luckhart and K.~B. Irani, ``An algorithmic solution of n-person games.'' in
  \emph{AAAI}, vol.~86, 1986, p. 158–162.

\bibitem{Korf_1991}
R.~E. Korf, ``Multi-player alpha-beta pruning,'' \emph{Artificial
  Intelligence}, vol.~48, no.~1, p. 99–111, 1991.

\bibitem{Carmel_Markovitch_1997}
D.~Carmel and S.~Markovitch, \emph{Pruning algorithms for multi-model adversary
  search}, 1997.

\bibitem{Sturtevant_Korf_2000}
N.~R. Sturtevant and R.~E. Korf, ``On pruning techniques for multi-player
  games,'' \emph{AAAI/IAAI}, vol.~49, p. 201–207, 2000.

\bibitem{Schadd_Winands_2011}
M.~P. Schadd and M.~H. Winands, ``Best reply search for multiplayer games,''
  \emph{IEEE Transactions on Computational Intelligence and AI in Games},
  vol.~3, no.~1, p. 57–66, 2011.

\bibitem{Lorenz_Tscheuschner_2006}
\BIBentryALTinterwordspacing
U.~Lorenz and T.~Tscheuschner, \emph{Player Modeling, Search Algorithms and
  Strategies in Multi-player Games}.\hskip 1em plus 0.5em minus 0.4em\relax
  Springer Berlin Heidelberg, 2006, vol. 4250, p. 210–224. [Online].
  Available: \url{http://link.springer.com/10.1007/11922155_16}
\BIBentrySTDinterwordspacing

\bibitem{Baier_Kaisers_2020}
H.~Baier and M.~Kaisers, ``Guiding multiplayer mcts by focusing on yourself,''
  in \emph{IEEE Conference on Games (CoG)}, 2020.

\bibitem{Sturtevant_2008}
N.~Sturtevant, ``An analysis of {UCT} in multi-player games,'' \emph{ICGA
  Journal}, p.~14, 2008.

\bibitem{Sturtevant_Zinkevich_Bowling_2006}
N.~Sturtevant, M.~Zinkevich, and M.~Bowling, ``Prob-max\^n: Playing n-player
  games with opponent models,'' in \emph{AAAI}, vol.~6, 2006, p. 1057–1063.

\bibitem{Donkers_2003}
H.~H. L.~M. Donkers, ``Nosce hostem: Searching with opponent models,'' 2003.

\bibitem{Carmel_Markovitch_1996}
D.~Carmel and S.~Markovitch, ``Incorporating opponent models into adversary
  search,'' in \emph{AAAI/IAAI, Vol. 1}, 1996, p. 120–125.

\bibitem{Coulom_2006}
R.~Coulom, ``Efficient selectivity and backup operators in monte-carlo tree
  search,'' in \emph{International conference on computers and games}.\hskip
  1em plus 0.5em minus 0.4em\relax Springer, 2006, p. 72–83.

\bibitem{Chaslot_De_Jong_Saito_Uiterwijk_2006}
G.~Chaslot, S.~De~Jong, J.-T. Saito, and J.~Uiterwijk, ``Monte-carlo tree
  search in production management problems,'' in \emph{Proceedings of the 18th
  BeNeLux Conference on Artificial Intelligence}, 2006, p. 91–98.

\bibitem{Browne_MCTS_2012}
C.~B. Browne, E.~Powley, D.~Whitehouse, S.~M. Lucas, P.~I. Cowling,
  P.~Rohlfshagen, S.~Tavener, D.~Perez, S.~Samothrakis, and S.~Colton, ``A
  survey of monte carlo tree search methods,'' \emph{IEEE Transactions on
  Computational Intelligence and AI in Games}, vol.~4, no.~1, p. 1–43, Mar
  2012.

\bibitem{UCT_2006}
L.~Kocsis and C.~Szepesvári, ``Bandit based monte-carlo planning,'' in
  \emph{European conference on machine learning}.\hskip 1em plus 0.5em minus
  0.4em\relax Springer, 2006, p. 282–293.

\bibitem{AlphaGo_2017}
D.~Silver, J.~Schrittwieser, K.~Simonyan, I.~Antonoglou, A.~Huang, A.~Guez,
  T.~Hubert, L.~Baker, M.~Lai, A.~Bolton, and et~al., ``Mastering the game of
  go without human knowledge,'' \emph{Nature}, vol. 550, no. 7676, p.
  354–359, Oct 2017.

\bibitem{Lisy_2014}
V.~Lisy, ``Alternative selection functions for information set monte carlo tree
  search,'' \emph{Acta Polytechnica}, vol.~54, no.~5, p. 333–340, 2014.

\bibitem{Cowling_Powley_Whitehouse_2012}
P.~I. Cowling, E.~J. Powley, and D.~Whitehouse, ``Information set monte carlo
  tree search,'' \emph{IEEE Transactions on Computational Intelligence and AI
  in Games}, vol.~4, no.~2, p. 120–143, Jun 2012.

\bibitem{Perez_Dieskau_Hunermund_Mostaghim_Lucas_2015}
\BIBentryALTinterwordspacing
D.~Perez~Liebana, J.~Dieskau, M.~Hunermund, S.~Mostaghim, and S.~Lucas, ``Open
  loop search for general video game playing,'' in \emph{Proceedings of the
  2015 on Genetic and Evolutionary Computation Conference - GECCO ’15}.\hskip
  1em plus 0.5em minus 0.4em\relax ACM Press, 2015, p. 337–344. [Online].
  Available: \url{http://dl.acm.org/citation.cfm?doid=2739480.2754811}
\BIBentrySTDinterwordspacing

\bibitem{Lucas_Liu_Perez-Liebana_2018}
\BIBentryALTinterwordspacing
S.~M. Lucas, J.~Liu, and D.~Perez-Liebana, ``The n-tuple bandit evolutionary
  algorithm for game agent optimisation,'' \emph{arXiv:1802.05991 [cs]}, Feb
  2018, arXiv: 1802.05991. [Online]. Available:
  \url{http://arxiv.org/abs/1802.05991}
\BIBentrySTDinterwordspacing

\bibitem{Lucas_Liu_Bravi_Gaina_Woodward_Volz_Perez-Liebana_2019}
S.~M. Lucas, J.~Liu, I.~Bravi, R.~D. Gaina, J.~Woodward, V.~Volz, and
  D.~Perez-Liebana, ``Efficient evolutionary methods for game agent
  optimisation: Model-based is best,'' \emph{arXiv preprint arXiv:1901.00723},
  2019.

\bibitem{van2009dimensionality}
L.~Van Der~Maaten, E.~Postma, J.~Van~den Herik \emph{et~al.}, ``Dimensionality
  reduction: a comparative,'' \emph{J Mach Learn Res}, vol.~10, no. 66-71,
  p.~13, 2009.

\bibitem{hotelling1933analysis}
H.~Hotelling, ``Analysis of a complex of statistical variables into principal
  components.'' \emph{Journal of educational psychology}, vol.~24, no.~6, p.
  417, 1933.

\bibitem{Maaten_Hinton_2008}
L.~v.~d. Maaten and G.~Hinton, ``Visualizing data using t-sne,'' \emph{Journal
  of machine learning research}, vol.~9, no. Nov, p. 2579–2605, 2008.

\bibitem{hayton2004factor}
J.~C. Hayton, D.~G. Allen, and V.~Scarpello, ``Factor retention decisions in
  exploratory factor analysis: A tutorial on parallel analysis,''
  \emph{Organizational research methods}, vol.~7, no.~2, pp. 191--205, 2004.

\bibitem{barber2012}
D.~Barber, \emph{Bayesian reasoning and machine learning}.\hskip 1em plus 0.5em
  minus 0.4em\relax Cambridge University Press, 2012.

\bibitem{Goodman_2021}
J.~Goodman, D.~Perez-Liebana, and S.~Lucas, ``Fingerprinting tabletop games,''
  in \emph{2021 IEEE Conference on Games (CoG)}, 2021.

\bibitem{Kim_Ashlock_2017}
E.-Y. Kim and D.~Ashlock, ``Changing resources available to game playing
  agents: Another relevant design factor in agent experiments,'' \emph{IEEE
  Transactions on Computational Intelligence and AI in Games}, vol.~9, no.~4,
  p. 321–332, 2017.

\bibitem{Jansen_1992}
P.~J. Jansen, \emph{Using Knowledge about the Opponent in Game-Tree Search.},
  1992.

\bibitem{Ramanujan_Sabharwal_Selman_2010}
R.~Ramanujan, A.~Sabharwal, and B.~Selman, ``On adversarial search spaces and
  sampling-based planning.'' in \emph{ICAPS}, vol.~10, 2010, p. 242–245.

\bibitem{Finnsson_Bjornsson_2011}
H.~Finnsson and Y.~Björnsson, ``Game-tree properties and {MCTS} performance,''
  in \emph{IJCAI}, vol.~11, 2011, p. 23–30.

\bibitem{Long_Sturtevant_Buro_Furtak_2010}
J.~R. Long, N.~R. Sturtevant, M.~Buro, and T.~Furtak, ``Understanding the
  success of perfect information monte carlo sampling in game tree search.'' in
  \emph{AAAI}, 2010.

\bibitem{Horn_Volz_Perez-Liebana_Preuss_2016}
\BIBentryALTinterwordspacing
H.~Horn, V.~Volz, D.~Perez-Liebana, and M.~Preuss, ``{MCTS/EA} hybrid {GVGAI}
  players and game difficulty estimation,'' in \emph{2016 IEEE Conference on
  Computational Intelligence and Games (CIG)}.\hskip 1em plus 0.5em minus
  0.4em\relax IEEE, Sep 2016, p. 1–8. [Online]. Available:
  \url{http://ieeexplore.ieee.org/document/7860384/}
\BIBentrySTDinterwordspacing

\bibitem{Togelius_Nelson_Liapis_2014}
J.~Togelius, M.~J. Nelson, and A.~Liapis, ``Characteristics of generatable
  games,'' 2014.

\bibitem{Browne_Maire_2010}
C.~Browne and F.~Maire, ``Evolutionary game design,'' vol.~2, p. 1–16, Mar
  2010.

\bibitem{Nelson_2011}
M.~J. Nelson, ``Game metrics without players: Strategies for understanding game
  artifacts,'' in \emph{Workshops at the Seventh Artificial Intelligence and
  Interactive Digital Entertainment Conference}, 2011.

\bibitem{volz20154}
V.~Volz, D.~Ashlock, and S.~Colton, ``4.18 gameplay evaluation measures,''
  \emph{Artificial and Computational Intelligence in Games: AI-Driven Game
  Design}, p. 122, 2015.

\bibitem{Guerrero-Romero_Louis_Perez-Liebana_2017}
\BIBentryALTinterwordspacing
C.~Guerrero-Romero, A.~Louis, and D.~Perez-Liebana, ``Beyond playing to win:
  Diversifying heuristics for gvgai.''\hskip 1em plus 0.5em minus 0.4em\relax
  IEEE, Aug 2017, p. 118–125. [Online]. Available:
  \url{http://ieeexplore.ieee.org/document/8080424/}
\BIBentrySTDinterwordspacing

\bibitem{Stephenson_Anderson_Khalifa_Levine_Renz_Togelius_Salge_2020}
M.~Stephenson, D.~Anderson, A.~Khalifa, J.~Levine, J.~Renz, J.~Togelius, and
  C.~Salge, ``A continuous information gain measure to find the most
  discriminatory problems for {AI} benchmarking,'' in \emph{2020 IEEE Congress
  on Evolutionary Computation (CEC)}, 2020.

\bibitem{soemers2021deep}
D.~J. Soemers, V.~Mella, C.~Browne, and O.~Teytaud, ``Deep learning for general
  game playing with ludii and polygames,'' \emph{arXiv preprint
  arXiv:2101.09562}, 2021.

\bibitem{Gaina_Liu_Lucas_Perez-Liebana_2017}
\BIBentryALTinterwordspacing
R.~D. Gaina, J.~Liu, S.~M. Lucas, and D.~Perez-Liebana, ``Analysis of vanilla
  rolling horizon evolution parameters in general video game playing,''
  \emph{arXiv:1704.07075 [cs]}, Apr 2017, arXiv: 1704.07075. [Online].
  Available: \url{http://arxiv.org/abs/1704.07075}
\BIBentrySTDinterwordspacing

\bibitem{Omidshafiei_et_2020}
S.~Omidshafiei, K.~Tuyls, W.~M. Czarnecki, F.~C. Santos, M.~Rowland, J.~Connor,
  D.~Hennes, P.~Muller, J.~Pérolat, B.~D. Vylder, and et~al., ``Navigating the
  landscape of multiplayer games,'' \emph{Nature Communications}, vol.~11,
  no.~11, p. 5603, Nov 2020.

\bibitem{SpinningTops_2020}
W.~M. Czarnecki, G.~Gidel, B.~Tracey, K.~Tuyls, S.~Omidshafiei, D.~Balduzzi,
  and M.~Jaderberg, ``Real world games look like spinning tops,'' \emph{arXiv
  preprint arXiv:2004.09468}, 2020.

\end{thebibliography}

\end{document}